\documentclass[11pt]{article}
\usepackage{eacl2017}
\usepackage{times}
\usepackage{url}
\usepackage{latexsym}

\eaclfinalcopy 

\usepackage[utf8]{inputenc}
\usepackage[T1]{fontenc}

\usepackage{multirow}

\usepackage{siunitx}
\usepackage{tikz}
\usepackage{pgfplots}

\usepackage{color}

\newcommand{\lingeval}{LingEval97}
\newcommand{\lingevalsize}{\num{97000} }

\makeatletter
\newcommand{\@BIBLABEL}{\@emptybiblabel}
\newcommand{\@emptybiblabel}[1]{}
\makeatother

\title{How Grammatical is Character-level Neural Machine Translation?\\ Assessing MT Quality with Contrastive Translation Pairs}

\author{
Rico Sennrich\\
School of Informatics, University of Edinburgh\\
{\tt \{rico.sennrich\}@ed.ac.uk}
}

\date{}

\begin{document}
\maketitle
\begin{abstract}
Analysing translation quality in regards to specific linguistic phenomena has historically been difficult and time-consuming.
Neural machine translation has the attractive property that it can produce scores for arbitrary translations, and we propose a novel method to assess how well NMT systems model specific linguistic phenomena such as agreement over long distances, the production of novel words, and the faithful translation of polarity.
The core idea is that we measure whether a reference translation is more probable under a NMT model than a contrastive translation which introduces a specific type of error.
We present \lingeval\footnote{Test set and evaluation script are available at \url{https://github.com/rsennrich/lingeval97}}, a large-scale data set of \lingevalsize contrastive translation pairs based on the WMT English$\to$German translation task, with errors automatically created with simple rules.
We report results for a number of systems, and find that recently introduced character-level NMT systems perform better at transliteration than models with byte-pair encoding (BPE) segmentation, but perform more poorly at morphosyntactic agreement, and translating discontiguous units of meaning.
\end{abstract}

\section{Introduction}

It has historically been difficult to analyse how well a machine translation system can learn specific linguistic phenomena.
Automatic metrics such as {\sc Bleu} \cite{Papineni2002} provide no linguistic insight, and automatic error analysis \cite{zora62156,DBLP:journals/pbml/Popovic11} is also relatively coarse-grained.
A concrete research question that has been unanswered so far is whether character-level decoders for neural machine translation \cite{DBLP:journals/corr/ChungCB16,2016arXiv161003017L} can generate coherent and grammatical sentences. \newcite{DBLP:journals/corr/ChungCB16} argue that the answer is yes, because {\sc Bleu} on long sentences is similar to a baseline with longer subword units created via byte-pair encoding (BPE) \cite{sennrich-haddow-birch:2016:WMT}, but {\sc Bleu}, being based on precision of short n-grams, is an unsuitable metric to measure the global coherence or grammaticality of a sentence.
To allow for a more nuanced analysis of different machine translation systems, we introduce a novel method to assess neural machine translation that can capture specific error categories in an automatic, reproducible fashion.

Neural machine translation \cite{kalchbrenner13emnlp,DBLP:conf/nips/SutskeverVL14,DBLP:journals/corr/BahdanauCB14} opens up new opportunities for automatic analysis because it can assign scores to arbitrary sentence 
pairs, in contrast to phrase-based systems, which are often unable to reach the reference translation.
We exploit this property for the automatic evaluation of specific aspects of translation by pairing a human reference translation with a contrastive example that is identical except for a specific error.
Models are tested as to whether they assign a higher probability to the reference translation than to the contrastive example.

\begin{table*}
\centering
\scriptsize
\begin{tabular}{c|lll}
category & English & German (correct) & German (contrastive) \\
\hline
NP agreement & [...] \textbf{of the} American \textbf{Congress} & [...] \textbf{des} amerikanischen \textbf{Kongresses} & * [...] \textbf{der} amerikanischen \textbf{Kongresses}\\
\hline
subject-verb agr. & [...] that the \textbf{plan} \textbf{will} be approved & [...], dass der \textbf{Plan} verabschiedet \textbf{wird} & * [...], dass der \textbf{Plan} verabschiedet \textbf{werden} \\ 
\hline
separable verb particle & he is \textbf{resting} & er \textbf{ruht} sich \textbf{aus} & * er \textbf{ruht} sich \textbf{an}\\
\hline
polarity & the timing [...] is \textbf{uncertain} & das Timing [...] ist \textbf{unsicher} & das Timing [..] ist \textbf{sicher}\\
\hline
transliteration & Mr. \textbf{Ensign's} office & Senator \textbf{Ensigns} Büro & Senator \textbf{Enisgns} Büro\\
\end{tabular}
\caption{Example contrastive translations pair for each error category.}
\label{examples}
\end{table*}

A similar method of assessment has previously been used for monolingual language models \cite{sennrichhaddow15,2016arXiv161101368L}, and we apply it to the task of machine translation.
We present a large-scale test set of English$\to$German contrastive translation pairs that allows for the automatic, quantitative analysis of a number of linguistically interesting phenomena that have previously been found to be challenging for machine translation, including agreement over long distances \cite{koehn-hoang:07,williams11}, 
discontiguous verb-particle constructions \cite{Niessen00improvingsmt,LOICIGA16.628},
generalization to unseen words (specifically, transliteration of names \cite{DBLP:conf/eacl/DurraniSHK14}), and ensuring that polarity is maintained \cite{wetzel-bond:2012:SSST-6,chen-zhu:2014:EACL,fancellu-webber:2015:ExProM}.

We report results for neural machine translation systems with different choice of subword unit, identifying strengths and weaknesses of recently-proposed models.

\section{Contrastive Translation Pairs}

We create a test set of contrastive translation pairs from the EN$\to$DE test sets from the WMT shared translation task.\footnote{\url{http://www.statmt.org/wmt16/}}
Each contrastive translation pair consists of a correct reference translation, and a contrastive example that has been minimally modified to introduce one translation error.
We define the accuracy of a model as the number of times it assigns a higher score to the reference translation than to the contrastive one, relative to the total number of predictions.
We have chosen a number of phenomena that are known to be challenging for the automatic translation from English to German.

\begin{enumerate}
\item noun phrase agreement: German determiners must agree with their head noun in case, number, and gender. We randomly change the gender of a singular definite determiner to introduce an agreement error.
\item subject-verb agreement: subjects and verbs must agree with one another in grammatical number and person. We swap the grammatical number of a verb to introduce an agreement error.
\item separable verb particle: verbs and their separable prefix often form a discontiguous semantic unit.
We replace a separable verb particle with one that has never been observed with the verb in the training data.
\item polarity: arguably, polarity errors are under-measured the most by string-based MT metrics, since a single word/morpheme can reverse the meaning of a translation. We reverse polarity by deleting/inserting the negation particle \emph{nicht} ('not'), swapping the determiner \emph{ein} ('a') and its negative counterpart \emph{kein} ('no'), or deleting/inserting the negation prefix \emph{un-}.
\item transliteration: subword-level models should be able to copy or transliterate names, even unseen ones. For names that were unseen in the training data, we swap two adjacent characters.
\end{enumerate}

Table~\ref{examples} shows examples for each error type.
Most are motivated by frequent translation errors; for EN$\to$DE, source and target script are the same, so technically, we do not perform transliteration.
Since transliteration of names and copying them is handled the same way by the encoder-decoder networks that we tested, we consider this error type a useful proxy to test the models' transliteration capability.

All errors are introduced automatically, relying on statistics from the training corpus, a syntactic analysis with ParZu \cite{sennrich13c}, and a finite-state morphology \cite{schmid2004,sennrich14} to identify the relevant constructions and introduce errors.
For contrastive pairs with agreement errors, we also annotate the distance between the words.
For translation errors where we want to assess generalization to rare words (all except negation particles), we also provide the training set frequency of the word involved in the error (in case of multiple words, we report the lower frequency).

The automatic processing has limitations, and we opt for a high-precision approach -- for instance, we only change the gender of determiners where case and number are unambiguous, so that we can produce maximally difficult errors.\footnote{If we mistakenly introduce a case error, this makes it easier to spot from local context.}
We expect that parsing errors will not invalidate the contrastive examples -- correctly identifying the subject will affect the distance annotation, but changing the number of the verb should always introduce an error.\footnote{Because of syncretism in German, there are cases where changing the inflection of one word does not cause disfluency, but merely changes the meaning. While a language model may deem both variants correct, a translation model should prefer the translation with the correct meaning.}
Still, we report ceiling scores achievable by humans to account for the possibility that a generated error is not actually an error.
We estimate the human ceiling by trying to select the correct variant for 20 contrastive translation pairs per category where our best system fails.
The ceiling is below 100\% because of errors in the reference translation,
and cases that were undecidable by a human annotator (such as the gender of \emph{the 20-year-old}).\footnote{We mark all undecidable cases as wrong, and could perform better with random guessing.}

From the \num{22191} sentences in the original newstest20** sets, we create approximately \lingevalsize contrastive translation pairs.

\begin{table}
\small
\centering
\setlength\tabcolsep{5pt}
\begin{tabular}{l|c|c|c}
& BPE--BPE & BPE--char & char--char \\
\hline
source vocab & 83,227 & 24,440 & 304 \\
target vocab & 91,000 & 302 & 302 \\
source emb. & 512 & 512 & 128 \\
source conv. & - & - & \scriptsize{\cite{2016arXiv161003017L}} \\
target emb. & 512 & 512 & 512 \\
encoder & gru & gru & gru \\
encoder size &  1024 & 512 & 512 \\
decoder & gru\_cond & \multicolumn{2}{c}{two\_layer\_gru\_decoder}\\
decoder size &  1024 & 1024 & 1024 \\
minibatch size & 128 & 128 & 64 \\
optmizer & adam & adam & adam \\
learning rate & 0.0001 & 0.0001 & 0.0001 \\
beam size & 12 & 20 & 20 \\
\hline
training time & $\approx$ 1 week & $\approx$ 2 weeks & $\approx$ 2 weeks\\
(minibatches) & 240,000 & 510,000 & 540,000 \\
\end{tabular}
\caption{NMT hyperparameters. `decoder' refers to function implemented in Nematus (for BPE-to-BPE) and dl4mt-c2c (for *-to-char).}
\label{hyperparameters}
\end{table}

\section{Evaluation}

In the evaluation section, our focus is on establishing baselines on the test set,
and investigating the following research questions:

\begin{itemize}
\item how well do different subword-level models process unseen words, specifically names?
\item sequence-length is increased in character-level models, compared to word-level or BPE-level models.
Does this have a negative effect on grammaticality?
\end{itemize}

\subsection{Data and Methods}

We train NMT systems with training data from the WMT 15 shared translation task EN$\to$DE.
We train three systems with different text representations on the parallel part of the training set:

\begin{itemize}
\item BPE-to-BPE \cite{sennrich-haddow-birch:2016:WMT}
\item BPE-to-char \cite{DBLP:journals/corr/ChungCB16}
\item char-to-char \cite{2016arXiv161003017L}
\end{itemize}

We use the implementations released by the respective authors, Nematus\footnote{\scriptsize{\url{https://github.com/rsennrich/nematus}}} for BPE-to-BPE,
and dl4mt-c2c\footnote{\scriptsize{\url{https://github.com/nyu-dl/dl4mt-c2c}}} for BPE-to-char and char-to-char.
dl4mt-c2c also provides preprocessed training data, which we use for comparability.

Both tools are forks of the dl4mt tutorial\footnote{\scriptsize{\url{https://github.com/nyu-dl/dl4mt-tutorial}}}, so the implementation differences are minimal except for those pertaining to the text representation.
We report hyperparameters in Table \ref{hyperparameters}.
They correspond to those used by \newcite{2016arXiv161003017L} for BPE-to-char and char-to-char;
for BPE-to-BPE, we also adopt some hyperparameters from \newcite{DBLP:journals/corr/SennrichHB15}, most importantly, we extract a joint BPE vocabulary of size \num{89500} from the parallel corpus.
We trained the BPE-to-BPE system for one week, following \newcite{sennrich-haddow-birch:2016:WMT}, and the *-to-char systems for two weeks, following \newcite{2016arXiv161003017L}, on a single Titan X GPU.
For both translating and scoring, we normalize probabilities by length (the number of symbols on the target side).

\begin{table}
\centering
\footnotesize
\setlength{\tabcolsep}{4pt}
\begin{tabular}{l|c|c|c}
system & 2014 & 2015 & 2016\\
(test set and size$\rightarrow$) & \scriptsize{3003} & \scriptsize{2169} & \scriptsize{2999} \\
\hline
BPE-to-BPE & 20.1 \scriptsize{(21.0)}& 23.2 \scriptsize{(23.0)} & 26.7 \scriptsize{(26.5)} \\ 
BPE-to-char & 19.4 \scriptsize{(20.5)} & 22.7 \scriptsize{(22.6)} & 26.0 \scriptsize{(25.9)} \\ 
char-to-char & 19.7 \scriptsize{(20.7)}& 22.9 \scriptsize{(22.7)} & 26.2 \scriptsize{(26.1)}\\ 
\hline
\scriptsize{\cite{sennrich-haddow-birch:2016:WMT}} & 25.4 \scriptsize{(26.5)} & 28.1 \scriptsize{(28.3)} & 34.2 \scriptsize{(34.2)}\\
\end{tabular}
\caption{Case-sensitive {\sc Bleu} scores (EN-DE) on WMT newstest. We report scores with detokenized NIST {\sc Bleu} ({\tt mteval-v13a.pl}), and in brackets, tokenized {\sc Bleu} with {\tt multi-bleu.perl}.
}
\label{results-bleu}
\end{table}

\begin{table*}
\centering
\footnotesize
\begin{tabular}{l|c|c|c|c|c|c}
& \multicolumn{2}{c|}{agreement} & & \multicolumn{2}{c|}{polarity (negation)} &  \\
system & noun phrase & subject-verb & verb particle & insertion & deletion & transliteration \\
(category and size$\rightarrow$) & 21813 & 35105 & 2450 & 22760 & 4043 & 3490 \\
\hline
BPE-to-BPE & \textbf{95.6} & \textbf{93.4} & \textbf{91.1} & 97.9 & \textbf{91.5} & 96.1 \\ 
BPE-to-char & 93.9 & 91.2 & 88.0 & \textbf{98.5} & 88.4 & \textbf{98.6} \\ 
char-to-char & 93.9 & 91.5 & 86.7 & \textbf{98.5} & 89.3 & \textbf{98.3} \\ 
\hline
\cite{sennrich-haddow-birch:2016:WMT} & 98.7 & 96.6 & 96.1 & 98.7 & 92.7 & 96.4 \\ 
\hline
human & 99.4 & 99.8 & 99.8 & 99.9 & 98.5 & 99.0 \\ 
\end{tabular}
\caption{Accuracy (in percent) of models on different categories of contrastive errors. Best single model result in bold (multiple bold results indicate that difference to best system is not statistically significant).}
\label{results}
\end{table*}

We also report results with the top-ranked system at WMT16 \cite{sennrich-haddow-birch:2016:WMT}, which is available online.\footnote{\scriptsize{\url{http://data.statmt.org/rsennrich/wmt16_systems/}}}
It is also a BPE-to-BPE system, but in contrast to the previous systems, it includes different preprocessing (including truecasing), other hyperparameters, additional monolingual training data, an ensemble of models, and bidirectional decoding.

\subsection{Results}

Firstly, we report case-sensitive {\sc Bleu} scores for all systems we trained for comparison to previous work.\footnote{Two commonly used {\sc Bleu} evaluation scripts,
the NIST {\sc Bleu} scorer {\tt mteval-v13a.pl} on detokenized text, and {\tt multi-bleu.perl} on tokenized text, give different results due to tokenization differences.
We here report both for comparison, but encourage the use of the NIST scorer, which is used by the WMT and IWSLT shared tasks, and allows for comparison of systems with different tokenizations.}
Results are shown in Table \ref{results-bleu}.
The results confirm that our systems are comparable to previously reported results \cite{sennrich-haddow-birch:2016:WMT,DBLP:journals/corr/ChungCB16},
and that performance of the three systems is relatively close in terms of {\sc Bleu}.
The metric does not provide any insight into the respective strengths and weaknesses of different text representations.

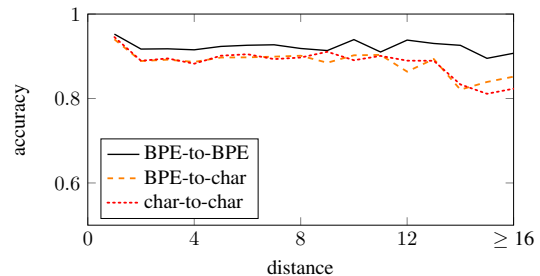
\begin{figure}
\centering
\begin{tikzpicture}[scale=0.7]
\pgfplotsset{major grid style={style=dotted,color=black!20}}

\pgfplotsset{
    overwrite last x tick label/.style={
        every x tick label/.append style={alias=lasttick},
        extra description/.append code={
            \fill [white] (lasttick.north west) ++(0pt,-\pgflinewidth) rectangle (lasttick.south east);         
            \node [anchor=base] at (lasttick.base) {#1};}
    },
    overwrite last x tick label/.default={$\ge$ 16}
}

\begin{axis}[xlabel=distance,
    xmin=0,
    xmax=16,
    ymin=0.5,
    ymax=1,
    ylabel=accuracy,
    xtick={0,4,8,12,16},
        overwrite last x tick label,
    legend pos = south west,
    y=8cm,
    x=0.5cm,
    legend style={
        /tikz/nodes={anchor=west}
        },
    mark size = 0.1,
    ]

    \addplot +[black, no markers, raw gnuplot, line width=0.15ex, id=bpe2bpe] gnuplot {plot 'plots/bpe2bpe.plot' };
    \addplot +[orange, no markers, raw gnuplot, dashed, line width=0.2ex, id=bpe2char] gnuplot {plot 'plots/bpe2char.plot'};
    \addplot +[red, no markers, raw gnuplot, dotted, line cap=round, line width=0.2ex, id=char2char] gnuplot {plot 'plots/char2char.plot' };

    \addlegendentry{BPE-to-BPE}
    \addlegendentry{BPE-to-char}
    \addlegendentry{char-to-char}

\end{axis}
\end{tikzpicture} 
\caption{Subject-verb agreement accuracy as a function of distance between subject and verb.}
\label{agreement-by-distance}
\end{figure}

Our main result is the assessment via contrastive translation pairs, shown in Table~\ref{results}.
We find that despite obtaining similar {\sc Bleu} scores, the models have learned different structures to a different degree.
The models with character decoder make fewer transliteration errors than the BPE-to-BPE model.
However, they perform more poorly on separable verb particles and agreement, especially as distance increases, as seen in Figure~\ref{agreement-by-distance}.
While accuracy for subject-verb agreement of adjacent words is similar across systems (95.2\%, 94.0\%, and 94.5\% for BPE-to-BPE, BPE-to-char, and char-to-char, respectively),
the gap widens for agreement between distant words -- for a distance of over 15 words, the accuracy is 90.7\%, 85.2\%, and 82.3\%, respectively.

\begin{table}
\centering
\tiny
\begin{tabular}{l|c|c|c|c|c|c}
& \multicolumn{3}{c|}{negation insertion} & \multicolumn{3}{c}{negation deletion} \\
system & \emph{nicht} & \emph{kein} & \emph{un-} & \emph{nicht} & \emph{kein} & \emph{un-} \\
(category and size $\rightarrow$) & 1297 & 10219 & 11244 & 2919 & 538 & 586 \\
\hline
BPE-to-BPE & \textbf{94.8} & \textbf{99.1} & 97.1 & \textbf{93.0} & \textbf{88.7} & \textbf{86.5} \\ 
BPE-to-char & 92.7 & \textbf{98.9} & \textbf{98.7} & 91.0 & 85.1 & 78.8 \\ 
char-to-char & 92.1 & \textbf{98.9} & \textbf{98.8} & 91.5 & 86.4 & 80.5 \\ 
\hline
\cite{sennrich-haddow-birch:2016:WMT} & 97.1 & 99.7 & 98.0 & 93.6 & 92.0 & 88.4 \\ 
\end{tabular}
\caption{Accuracy (in percent) of models on different categories of contrastive errors related to polarity. Best single model result in bold.}
\label{results-negation}
\end{table}

Polarity shifts between the source and target text are a well-known translation problem, and our analysis shows that the main type of error is the deletion of negation markers,
in line with with findings of previous studies \cite{fancellu-webber:2015:ExProM}.
We consider the relatively high number of errors related to polarity an important problem in machine translation, and hope that future work will try to improve upon our results, shown in more detail in Table~\ref{results-negation}.

We have commented that changing the grammatical number of the verb may change the meaning of the sentence instead of making it disfluent.
A common example is the German pronoun \emph{sie}, which is shared between the singular 'she', and the plural 'they'.
We keep separate statistics for this type of error ($n=2520$), and find that it is challenging for all models, with an accuracy of 87--87.2\% for single models, and 90\% by the WMT16 submission system.

\begin{table*}
\centering
\small
\setlength\tabcolsep{3pt}
\begin{tabular}{l|l|c}
system & sentence & cost\\
\hline
source & Since then we have only played in the Swedish \textbf{league which is} not the same level. \\ 
reference & Seitdem haben wir nur in der Schwedischen \textbf{Liga} gespielt, \textbf{die} nicht das gleiche Niveau \textbf{hat}. & 0.149\\
contrastive & Seitdem haben wir nur in der Schwedischen \textbf{Liga} gespielt, \textbf{die} nicht das gleiche Niveau \textbf{haben}. & 0.137\\
1-best & Seitdem haben wir nur in der schwedischen \textbf{Liga} gespielt, \textbf{die} nicht die gleiche Stufe \textbf{sind}. & 0.090\\
\hline
source & FriendsFest: the \textbf{comedy show that taught} us serious lessons about male friendship. \\ 
reference & FriendsFest: die \textbf{Comedy-Show, die} uns ernsthafte Lektionen über Männerfreundschaften \textbf{erteilt} & 0.276\\
contrastive & FriendsFest: die \textbf{Comedy-Show, die} uns ernsthafte Lektionen über Männerfreundschaften \textbf{erteilen} & 0.262\\
1-best & FriendsFest: die \textbf{Komödie} zeigt, \textbf{dass} uns ernsthafte Lehren aus männlichen Freundschaften & 0.129\\
\hline
source & Robert Lewandowski \textbf{had} the best opportunities in the first half. \\ 
reference & Die besten Gelegenheiten in Hälfte eins \textbf{hatte} Robert Lewandowski. & 0.551\\
contrastive & Die besten Gelegenheiten in Hälfte eins \textbf{hatten} Robert Lewandowski. & 0.507\\
1-best & Robert Lewandowski \textbf{hatte} in der ersten Hälfte die besten Möglichkeiten. & 0.046 \\
\end{tabular}
\caption{Examples where char-to-char model prefers contrastive translation (subject-verb agreement errors). 1-best translation can make error of same type (example 1), different type (translation of \emph{taught} is missing in example 2), or no error (example 3).}
\label{examples2}
\end{table*}

We conclude from our results that there is currently a trade-off between generalization to unseen words, for which character-level decoders perform best, and sentence-level grammaticality, for which we observe better results with larger subword units of the BPE segmentation.
We hope that our test set will help in developing and assessing architectures that aim to overcome this trade-off and perform best in respect to both morphology and syntax.

We encourage the use of contrastive translation pairs, and \lingeval, for future analysis, but here discuss some limitations.
The first one is by design: being focused on specific translation errors, the evaluation is not suitable as a global quality metric.
Also, the evaluation only compares the probability of two translations, a reference translation $T$ and a contrastive translation $T'$,
and makes no statement about the most probable translation $T^*$.
Even if a model correctly estimates that $p(T) > p(T')$, it is possible that $T^*$ will contain an error of the same type as $T'$.
And even if a model incorrectly estimates that $p(T) < p(T')$, it may produce a correct translation $T^*$.
Despite these limitations, we argue that contrastive translation pairs are useful because they can easily be created to analyse any type of error in a way that is model-agnostic, automatic and reproducible.

Table~\ref{examples2} shows different examples of the where the contrastive translation is scored higher than the reference by the char-to-char model, and the corresponding 1-best translation.
In the first one, our method automatically recognizes an error that also appears in the 1-best translation.
In the second example, the 1-best translation is missing the verb.
Such cases could confound a human analysis of agreement errors, and we consider it an advantage of our method that it is not confounded by other errors in the 1-best translation.
In the third example, our method identifies an error, but the 1-best translation is correct.
We note that the German reference exhibits object fronting, but the 1-best output has the more common SVO word order.
While one could consider this instance a false positive, it can be important for an NMT model to properly score translations other than the 1-best, for instance for applications such as prefix-constrained MT \cite{P16-1007}.

\section{Conclusion}

We present \lingeval, a test set of \lingevalsize contrastive translation pairs for the assessment of neural machine translation systems.
By introducing specific translation errors to the contrastive translations, we gain valuable insight into the ability of state-of-the-art neural MT systems to handle several challenging linguistic phenomena.
A core finding is that recently proposed character-level decoders for neural machine translation outperform subword models at processing unknown names, but perform worse at modelling morphosyntactic agreement, where information needs to be carried over long distances.
We encourage the use of \lingeval{} to assess alternative architectures, such as hybrid word-character models \cite{2016arXiv160400788L}, or dilated convolutional networks \cite{kalchbrenner16}.
For the tested systems, the most challenging error type is the deletion of negation markers, and we hope that our test set will facilitate development and evaluation of models that try to improve in that respect.
Finally, the evaluation via contrastive translation pairs is a very flexible approach, and can be applied to new language pairs and error types.

\section*{Acknowledgments}

This project received funding from the European Union's Horizon 2020 research and innovation programme under grant agreements 645452 (QT21) and 688139 (SUMMA).

\bibliographystyle{eacl2017}
\bibliography{../bibliography.bib}

\end{document}